\documentclass[letterpaper, 10 pt, conference]{ieeeconf}
\IEEEoverridecommandlockouts
\usepackage{ulem}
\usepackage{amsmath, amsfonts, amssymb}
\usepackage{todonotes}
\usepackage{multirow}

\usepackage{cite}
\usepackage{array}
\newcolumntype{P}[1]{>{\centering\arraybackslash}p{#1}}
\usepackage{booktabs}
\usepackage{multicol}
\usepackage{graphicx}
\usepackage{censor}
\usepackage{hyperref}
\usepackage[belowskip=-4pt,aboveskip=0pt, font=small, labelfont=small]{caption}

\usepackage[pagewise]{lineno}
\usepackage{graphicx}
\graphicspath{{./figures/}}
\usepackage{caption}

\usepackage{subfig}
\usepackage{listings}
\usepackage{xcolor}

\definecolor{eclipseStrings}{RGB}{42,0.0,255}
\definecolor{eclipseKeywords}{RGB}{127,0,85}
\colorlet{numb}{magenta!60!black}

\lstdefinelanguage{json}{
    basicstyle=\normalfont\ttfamily,
    commentstyle=\color{eclipseStrings}, % style of comment
    stringstyle=\color{eclipseKeywords}, % style of strings
    numbers=left,
    numberstyle=\scriptsize,
    stepnumber=1,
    numbersep=8pt,
    showstringspaces=false,
    breaklines=true,
    frame=lines,
    string=[s]{"}{"},
    comment=[l]{:\ "},
    morecomment=[l]{:"},
    literate=
        *{0}{{{\color{numb}0}}}{1}
         {1}{{{\color{numb}1}}}{1}
         {2}{{{\color{numb}2}}}{1}
         {3}{{{\color{numb}3}}}{1}
         {4}{{{\color{numb}4}}}{1}
         {5}{{{\color{numb}5}}}{1}
         {6}{{{\color{numb}6}}}{1}
         {7}{{{\color{numb}7}}}{1}
         {8}{{{\color{numb}8}}}{1}
         {9}{{{\color{numb}9}}}{1}
}

\title{\LARGE \bf
Linear Delta Arrays for Compliant Dexterous Distributed Manipulation
}

\author{Sarvesh Patil, Tony Tao, Tess Hellebrekers, Oliver Kroemer, F. Zeynep Temel% <-this % stops a space
\thanks{S. Patil, T. Tao, O. Kroemer, and F. Z. Temel are with the Robotics Institute, Carnegie Mellon University, 
        Pittsburgh, PA 15213, USA
        {\tt\small \{sarveshp, longtao, okroemer, ztemel\} @andrew.cmu.edu}} \\
\thanks{Tess Hellebrekers is with Meta AI
        {\tt\small tessh@fb.com}} \\
\authorblockA{\url{https://iamlab-cmu.github.io/delta-arrays/}}}
\begin{document}
% \linenumbers % Uncomment this to enable line numbers in the peer review

\maketitle
\thispagestyle{empty}
\pagestyle{empty}

\begin{abstract}
%\sarvesh{Change abstract acc to Jacky's comment structure}

This paper presents a new type of distributed dexterous manipulator: delta arrays. Our delta array setup consists of 64 linearly-actuated delta robots with 3D-printed compliant linkages. Through the design of the individual delta robots, the modular array structure, and distributed communication and control, we study a wide range of in-plane and out-of-plane manipulations, as well as prehensile manipulations among subsets of neighboring delta robots. We also demonstrate dexterous manipulation capabilities of the delta array using reinforcement learning while leveraging compliance. Our evaluations show that the resulting 192 DoF compliant robot is capable of performing various coordinated distributed manipulations of a variety of objects, including translation, alignment, prehensile squeezing, lifting, and grasping.

\end{abstract}

\begin{keywords} % UPDATE
Multi-Robot Systems,  Soft Robot Applications,  Dexterous Manipulation %Additive Manufacturing, Soft Robot Applications, Flexible Robotics, Kinematics
%Telerobotics and Teleoperation, Physical Human-Robot Interaction,
% keywords, choose from \\ https://www.ieee-ras.org/publications/ra-l/keywords
\end{keywords}

\normalem
\section{Introduction}
The term \emph{dexterous manipulation} often invokes the image of a five-fingered hand delicately holding an object as a human would. However, robots are not restricted to human morphology. Imagine instead a surface covered in fingers. Each finger can move its fingertip in a small 3D workspace above its fixed base and interact with parts of objects that enter its workspace. The fingers can work together to shift, tilt, lift, block, and even pinch objects. The large number of fingers provides additional redundancy, with larger objects being manipulated by tens of fingers at a time. The distributed nature of the fingers also means that multiple objects can be easily manipulated in parallel in different regions of the surface. This type of system would thus represent a distributed dexterous manipulation paradigm. 

In this paper, we present an array of linear delta robots for the development of distributed dexterous manipulation strategies. Delta arrays consist of grids of small prismatic soft delta robots (3 degrees-of-freedom (DoF) each) that work together to manipulate objects. We propose a modular design for the delta arrays that consist of $2\times 2$ units (12 DoF each) with each unit having a standalone mechanical and electronic design. Each unit has its own processor and controllers, allowing for distributed computation with a central computer providing high-level commands. We also present a real hardware implementation of an $8\times 8$ array consisting of $16$ units and providing $192$ degrees of freedom. 

\begin{figure}
    \centering
    \captionsetup{width=.495\textwidth}
    \includegraphics[width=0.495\textwidth]{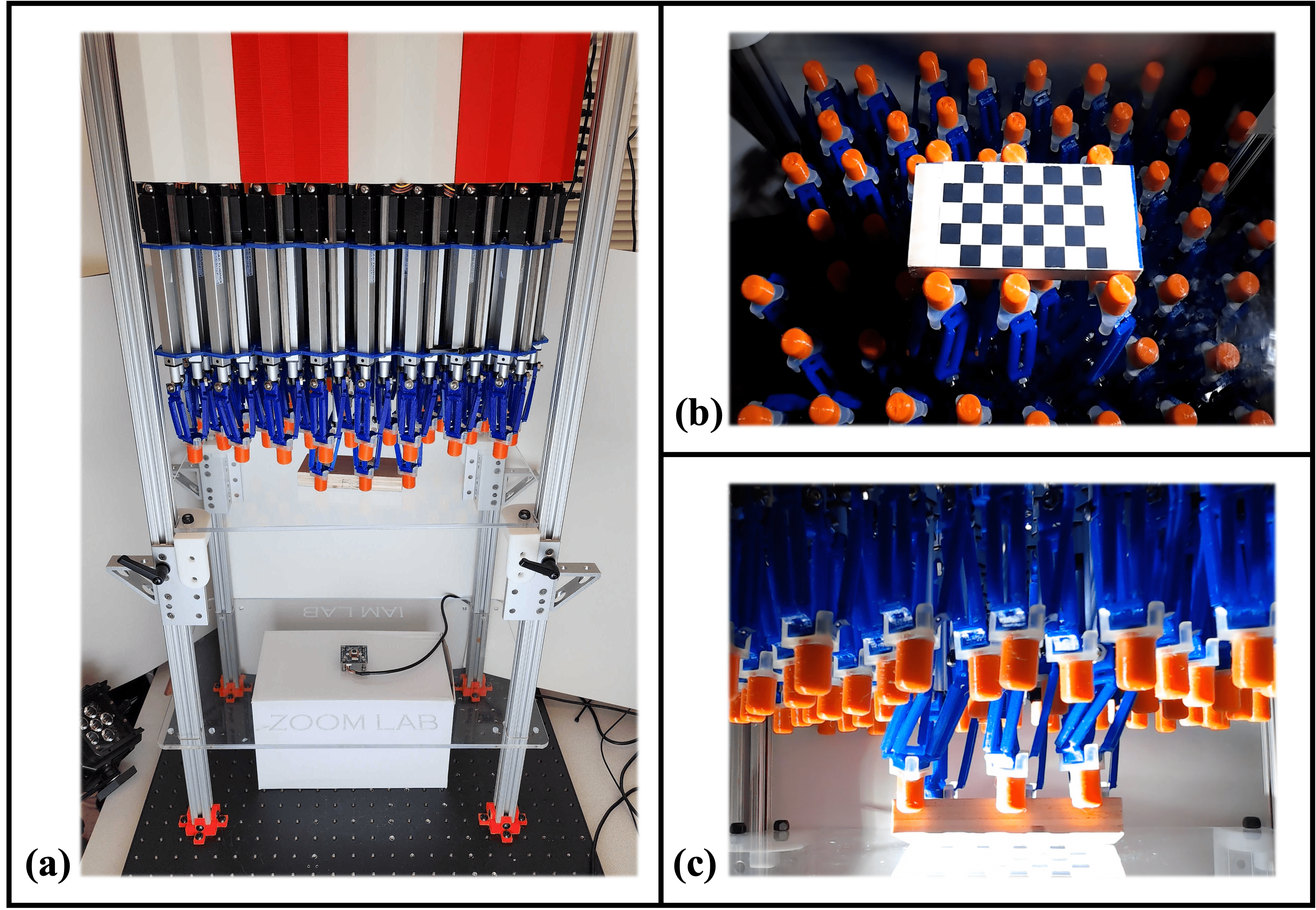}
    \caption{Delta array dexterous manipulation setup with robots facing down. (a) The setup consists of 64 delta robots actuated by linear motors and a camera. (b) Checkered board on the object used for pose estimation. (c) Distributed manipulation strategy for tilting wooden block. }
    \label{fig: upsideDown}
\end{figure}

Each compliant delta robot in the array is actuated by three linear actuators. These actuators are connected via parallel mechanisms to an end-effector platform. The platform and parallel linkages are 3D printed together out of thermoplastic polyurethane (TPU - 95A shore hardness) for easier assembly, compliant interactions, and low hysteresis under extreme deformations. The linear-actuator design allows for the delta robots to be packed closely together, in a hexagonal grid, and for their end-effectors to move outside of the footprint of the actuators. This allows the workspaces of neighboring deltas to overlap, and perform prehensile manipulations such as pinching between neighboring delta robots.

We present two modes of operating the delta array: 
\begin{itemize}
    \item Facing Down - Objects are placed on a plexiglass plane and manipulated from the top with a camera underneath the array for visual feedback (Fig. \ref{fig: upsideDown}). 
    \item Facing Up - Objects are placed on top of the array and manipulated from underneath or on the sides (Fig. \ref{fig: upsideUp}).
    
\end{itemize}

The delta array provides a basis for a wide range of different manipulation strategies. Similar to smart conveyors, delta arrays are capable of executing various planar transportation behaviors. Unlike smart conveyors, delta arrays need to use a finger gaiting approach, with coordinated making and breaking of contacts across delta robots, to shift objects across the array's workspace. This added complexity, however, means delta arrays can make better contact with objects that have non-planar surfaces. The additional flexibility and non-planar motions of the deltas also allow for a number of other strategies that require out-of-plane motions in 3D workspaces. These strategies contrast with traditional manipulators in the sense that the end-effectors are static as opposed to being mounted on a robotic arm, and rely more on cooperation between multiple agents to accomplish tasks.

In the facing-up mode, the variable height allows for the rolling and tilting of objects on the array surface. Fingers can be raised to create fixture-like structures for aligning objects. The lateral motions allow the deltas to grasp and pinch objects of various sizes across the array. Although many of these strategies have been individually supported by other distributed or dexterous manipulation systems, to the best of our knowledge, this is the first system that supports all of these strategies and thus the possibility of combining strategies, as well as learning new ones.  

\begin{figure}
    \centering
    \includegraphics[width=0.495\textwidth]{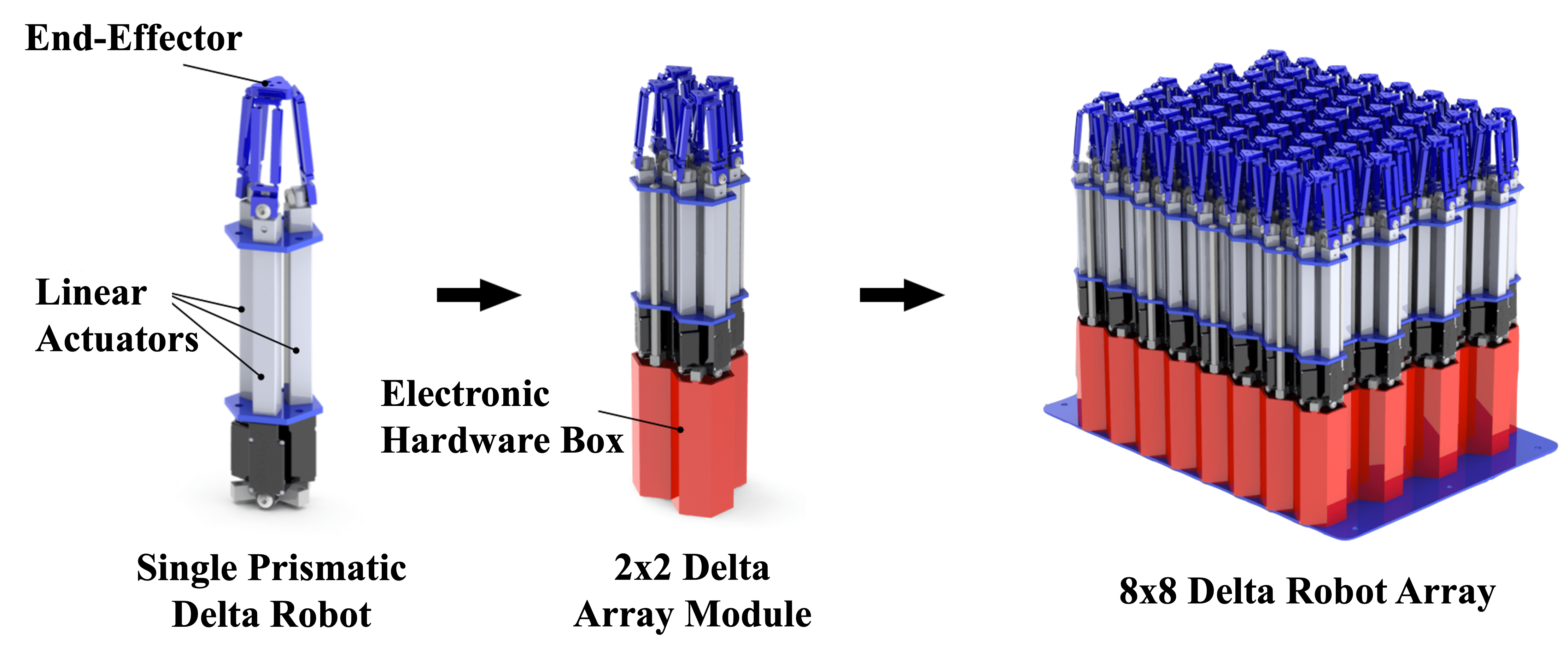}
    \caption{The modular design of the delta array. Each robot consists of three linear actuators and a 3D printed TPU end effector (left). Four robots organized in a $2\times2$ hexagonal grid form a module, which shares the electrical components (middle). 16 modules form the delta array (right).  %\zeynep{let me know what you want me to add/remove/change}}
    \label{fig:fig2}}
\end{figure} %can we do this? 

Implementing and controlling an array of delta robots presents a number of challenges. The design needs to be modular for easy construction, extension, and maintenance. The individual delta robots need to be robust and safe, but also precise and capable of supporting a wide range of manipulation strategies. The communication needs to be fast and scalable to minimize command latency throughout the array network. In the remainder of this paper, we will explain and discuss our design decisions in developing the delta arrays and how we tackle each of these systemic requirements.
Sections \ref{relatedwork}, \ref{design}, and \ref{array} describe the modular arrays in a bottom-up manner, while sections \ref{strats}, and \ref{dexmanip} present basic control strategies for manipulating objects with the arrays. The focus of this paper is on the design of the delta arrays and demonstrating its ability to execute a variety of distributed dexterous manipulation strategies. Developing more advanced and hybrid strategies is left to future work.  

% In Section \ref{relatedwork} we describe related work on delta robots and linear actuation array platforms. In Section \ref{design} we present the design of the individual delta robots for close packing and overlapping workspaces. In Section \ref{array} we explain how the arrays are constructed using standalone $2\times 2$ modules. Section \ref{strats} describes manually generated primitive manipulation strategies for transporting objects, while section \ref{dexmanip} demonstrates use of model free reinforcement learning (RL) techniques to learn dynamic motion primitives (DMPs) for dexterous manipulation tasks. 
% In Section \ref{exps} we present experiments of the real delta array for distributed dexterous manipulation. The focus of this paper is on the design of the delta arrays and demonstrating its ability to execute a variety of distributed dexterous manipulation strategies. Developing more advanced and hybrid strategies is left to future work.  

%The delta array is a distributed manipulation system that can achieve a wide range of manipulation tasks through cooperation and dynamic contact transferal. A proof of concept was recently shown by Thompson et al., in !CITE DELTA ARRAY SIMULATOR!. This work translates the design choices from simulation into a functional robotic system of delta robots arranged in an 8x8 hexagonal grid pattern. 

%. It has been inspired from the low-cost compliant gripper design 

\section{Related Work}\label{relatedwork}
%two approaches: (i) from hci perspective shape changing interfaces - dynamic tactile displays/surfaces and (ii) smart conveyors. anything else?
\subsection{Delta robots}

Delta robots were introduced by Clavel in 1990 and initially designed as a pick-and-place tool \cite{clavel1990device}.
Conventional delta robots have a fixed base and a moving stage that are always parallel to
each other. These platforms are connected by three kinematic chains with revolute and universal
joints. These chains are each driven by single-DoF actuators that are positioned at the fixed base. The motion is transmitted from the base arm to the moving stage by three parallelograms, which are the key to the delta robot’s functionality \cite{rey1999delta}.
In our recent work \cite{RSSdelta}, we presented a gripper based on two prismatic delta manipulators using 3D-printed parallelogram links presented in \cite{siciliano_characterization_2021}. Unlike traditional parallel jaw grippers, our robots have compliant end-effectors, which makes them modular and accessible. This 6-DoF system is able to perform dexterous manipulation tasks, such as aligning a pile of coins,  picking  up  a  card  from a  deck,  plucking  a  grape  off  of  a  stem,  and  rolling  dough.

\subsection{Robot hand and finger design for dexterous manipulation}
Current robot hands with fingers span a range of different designs and complexity. Basic two-fingered grippers often have a single DoF, while complex anthropomorphic hands will often have multiple DoF per finger \cite{Kim2021, shadow_openai}. Current designs use serial mechanisms for the
individual fingers, similar to human hands. However, the more dexterous designs either require relatively bulky motors to be placed in the fingers, where they significantly increase the inertia, or they are actuated by cable drives \cite{Andrychowicz2020}, which are subject to highly non-linear effects and temporal variations due to slack and friction along the
cables.

\subsection{Dynamic surfaces}
Dynamic surfaces have the potential to be used not only for object manipulation, but also as shape-changing interfaces. Distributed manipulation systems have many types, such as vibrating plates \cite{525555}, actively controlled arrays of air jets \cite{yim2000two}, planar micromechanical  actuator arrays \cite{bourbon1999toward}, and actuated workbenches using magnetic forces \cite{actuatedworkbench}. These dynamic surfaces with an actuator array are also widely used in interactive displays. However, prior works focus on the motion on a plane, rather than working on the motion in 3D space. An additional DoF on the normal surface allows the delta array to interact with objects while utilizing contacts in 3D space. 

\subsection{Distributed and dexterous manipulation primitives}
Dexterous manipulation using soft grippers is a challenging task due to stochasticity in the kinematics of the soft robot bodies. Various robust model-based control strategies use Lagrangian formulations to model the system \cite{euler_lagrange_soft_robot}\cite{euler_lagrange_2}. Another way is to use human demonstrations and Dynamic Motion Primitives (DMPs) to generate dynamically constrained trajectories for manipulation \cite{okamura}\cite{humanDemoDM_1}\cite{humanDemoDM_2}. However, using analytical methods for a multi-agent system can lead to excessively high demand in compute, and generating human demonstrations for such a high degree of freedom system is logistically infeasible. Thus, we deploy a model-free RL algorithm directly on the hardware to generate trajectories for pushing and tilting an object against other robots to demonstrate the dexterous manipulation capabilities of the delta arrays.

\section{Prismatic Delta Robots}\label{design}

% \begin{figure*}
%     \centering
%     \includegraphics[width=0.995\textwidth]{figures/assembly_sequence3.png}
%     \caption{The modular design of the delta array. Each robot consists of three linear actuators and a 3D printed TPU end effector (left). Four robots organized in a hexagonal grid form a module, which shares the electrical components (middle). 16 modules form the delta array (right).  %\zeynep{let me know what you want me to add/remove/change}}
%     \label{fig:fig2}}
% \end{figure*}

A delta array consists of multiple delta robots arranged in a planar grid structure. In this section, we explain the design of the individual delta robots. Each delta robot consists of three actuators connected by a parallel-bar linkage  end-effector platform, as shown in Fig.\ref{fig:fig2}.

\subsection{Actuators}
Delta robots are often designed with rotational actuators %, such as servo motors,  
that provide torque to individual links \cite{rey1999delta, sam_alvares_building_2021}. These designs provide rapid and precise movements at the end-effector, but at the cost of a wide robot base. Due to the excessive width, which would conflict with the goal of creating a closely packed array of delta robots, we utilize linear actuators that enable us to position each robot in close proximity to one another. The linear actuators (Actuonix) have a $10$ cm stroke length and internal potentiometers to provide analog feedback for position control. The end effector design is based on our previous work \cite{RSSdelta}.

\begin{figure}
    \centering
    \includegraphics[width=0.495\textwidth]{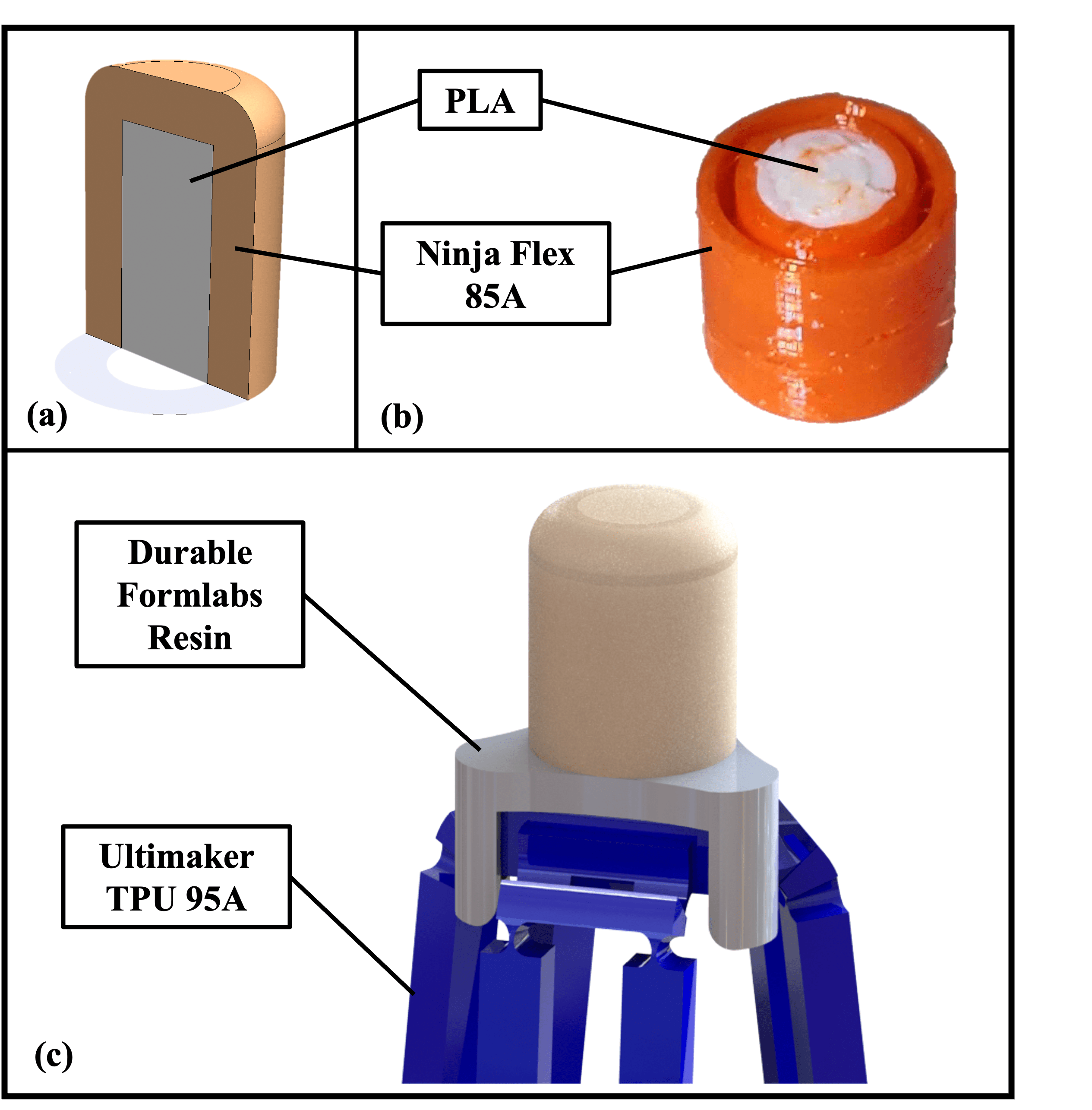}
    \caption{(a) Cross-section view of the fingertip assembly in CAD (b) Cross-section view of 3D printed fingertip shows air pocket created by $0\%$ infill (c) Attachment clip latches into underside of link using catch on end of the 3 prongs.
    \label{fig:finger_tips}}
\end{figure} %can we do this? 

\subsection{End Effector and Parallelogram Linkages}

The end-effector platform is connected to three actuators through a parallelogram link, which converts linear motions into precise 3D x-y-z motions at the end-effector while keeping the platform parallel to the base.  

The delta links are 3D printed as a single part with living hinges. Additional details can be found at \cite{siciliano_characterization_2021}. The delta links were printed using thermoplastic polyurethane (TPU) for its low Young's modulus. This compliance allows the robots to safely interact with objects, other robots, and reduce the wear and tear of the system. 

To perform dexterous manipulation, we design a fingertip that is inspired by the texture and feel of a human finger. The fingertip is attached as an end-effector to the delta platform using a reusable clip printed using durable formv3 resin for strength and flexibility. The fingertips were 3D printed using Polylactic acid (PLA) for the inner bone structure and NinjaFlex 85A for the outer skin which was fused together using a dual extrusion printer. Fig. \ref{fig:finger_tips} shows the cross-sectional structure of the design.We achieve this hollow structure by using $0\%$ infill for the NinjaFlex and thin, two-layer walls which results in an enclosed cavity that provides the surface compliance for the finger.

%The delta linkages benefit tremendously from the low Young's modulus that thermoplastic polyurethane (TPU) offers, especially when collisions occur between links in the array. In this work, we design delta links with 0.375mm hinges and 4.5mm thick beams 3D-printed with TPU. The flexible links allow the end effector to remain parallel to the base within a workspace radius of about 1.5cm, after which it starts tilting depending on the radial distance. The distance between each delta robot in the array compensates for the tilting, by mitigating any need for the robot to exceed the 1.5cm radial limit in workspace.

 \subsection{Delta Robot Workspace}
 
A key benefit of the prismatic delta design is that the workspace of the delta's end-effector extends beyond the triangular footprint of the three actuators. For our implementation, the horizontal distance between the centers of two actuators in a delta robot is 2cm, while the width of the workspace is approximately 6cm. To avoid excessive collisions between neighboring deltas, we restrict the horizontal workspace to a diameter of 3cm. The vertical workspace corresponds to the 10cm stroke length of the actuators.

The delta robots are operated within a workspace that is far away from their singularities. Ambiguities in the inverse kinematics can therefore be easily resolved to determine a suitable joint trajectory for a given desired end-effector trajectory. 

% \sarvesh{If time permits, compute max load capacity using Zilin's sensor and quantify weight in Kgs that the sys can manipulate at max.}

\section{Modular Array Structure}\label{array}
Sets of delta robots are arranged into hexagonal grids to create delta arrays. Rather than constructing an array out of single deltas, we instead developed a modular $2\times2$ array unit for four deltas. Each unit can be operated in a standalone manner and provides a shared set of electronics and microcontrollers. To create an $8\times8$ array, we simply place 16 of the modules in a $4\times4$ macro grid, and a central computer then communicates to all of the modules to create coordinated manipulation strategies. The $2\times2$ modules thus provide a modular and extendable basis for easily constructing arrays of different sizes and replacing parts as needed. Our $8\times8$ configuration allows the manipulation of objects of a range of sizes and demonstrates the potential of such arrays in dexterous tasks.

\subsection{$2\times2$ Delta Modules}

Each $2\times2$ module employs a hexagonal structure as shown in the middle image of Fig.\ref{fig:fig2}. The linear actuator bodies are held together using two laser-cut plexiglass plates, which are, in turn, supported by aluminum stand-offs. The stand-off configuration equally compresses the $12$ linear actuators from both sides forming a stable structure.

Each robot in the module is then secured through the base of the linear actuators using a 3D-printed connector made from PLA and then attached to a 3D-printed enclosure made of PLA. This hardware box houses the electronics needed to control the four deltas in that module and allows the module to be connected with a base plate that supports the array. The resulting $2\times2$ delta modules offer a balance between modularity and ease of maintenance.

% The base plate, which supports the modules, is constructed from a laser-cut sheet of plexiglass. The plate allows the $2\times2$ prismatic deltas to seamlessly connect and arranges them in a way that allows each robot's workspace to overlap.

For the face-down mode of operation, the setup is inverted and mounted onto pillars made of 80/20 aluminum extrusions---constrained at the top by the base plate and on the bottom by an optical breadboard. Between the array and the breadboard, a clear plexiglass platform is supported by a movable linear slide on each corner. Planar manipulations are performed ontop of this platform. These sliders allow the height of the platform to be manually adjusted depending on the size of the objects being manipulated. The transparent platform enables our vision pipeline to track the pose of objects from below as they are manipulated.

% The plate is mounted onto pillars made of 80/20 aluminum extrusions which enable us to position the plate in any orientation. The silicone washers used to mount the plate to the 80/20 damps vibrations prevent plexiglass from shattering under the heavy stress of the delta arrays.

%The large plate connects the modules from the base and facilitates mounting solutions in a variety of orientations for different tasks. The modular design allows for the delta array to be readily expanded or contracted to fit the needs of the application.

%Since we position 4 delta robots into each of these modules, only 16 modules need to be regulated rather than 64 individually controlled deltas on the management scale. This also allows us to utilize the powerful computation of a single state-of-the-art micro-controller to control 4 delta robots and enables us to package the robot in a small form factor.

\subsection{Electronics}

%To control the four deltas ($12$ actuators) in a module, we use Adafruit Feather M0 boards. These boards have a smaller form factor than the Arduino Megas used by our previous delta designs \cite{RSSdelta} and can thus be housed in the hardware enclosure box.

\begin{figure}
    \centering
    \captionsetup{width=\columnwidth}
    \includegraphics[width=\columnwidth]{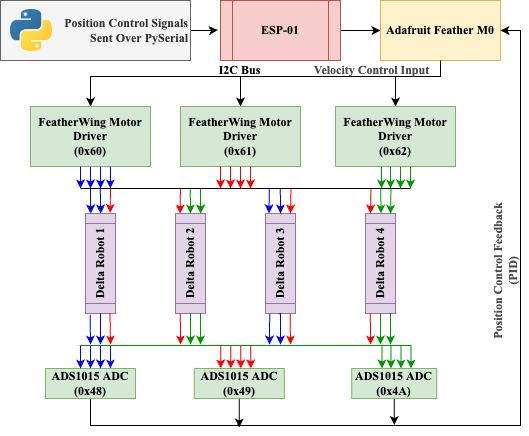}
    \caption{Communication flowchart for a $2\times2$ module. The control of three actuators of each robot in a four-robot module is accomplished by three motor drivers. Colored arrows show the distributed control framework between drivers, actuators, and ADCs.}
    \label{fig:fig4}
\end{figure}
Adafruit Feather M0 boards control four deltas ($12$ actuators) in a module and they are housed in the hardware enclosure box. Adafruit DC Motor/Stepper FeatherWing is used to control the velocity of the end effectors through PWM signals. We use an analog-to-digital converter (ADC) to couple the position feedback from the linear actuators. This also acts as a low-pass filter to eliminate high-frequency noise from the electromagnetic interference generated in the circuit for precise position control. We use a 12-bit ADC that resolves the 100 mm length, and we apply a low level PID control with a final precision of upto $0.3$ mm.

We use the $I^2C$ bus on the Feather M0 and distribute the data bus, clock, and a 12W power supply across three FeatherWing motor drivers using a custom electronic shield circuit. The compactness of the design allows us to maintain close proximity among all the delta modules. 

For perception, we mount a USB camera module on the bottom plane looking upwards. The camera tracks object poses with minimal occlusions while dexterous manipulation is being performed by the delta array from above. 

% We use the $I^2C$ bus on the Feather M0 to send and receive commands to the FeatherWing and the ADC. To control 12 linear actuators, we stack up three FeatherWing and ADC pairs using a custom circuit that takes care of the $I^2C$ address adders as well as power delivery to the FeatherWings' motor drivers. The FeatherWing $I^2C$ addresses are 0x60, 0x61, and 0x62, and those of the ADCs are 0x48, 0x49, and 0x4A respectively. A 12V 1A DC adapter delivers the power through a barrel jack, which is then distributed across the FeatherWings using the shield circuit. The electronics choices enable us to create the entire circuit with a form factor of about $50$mm $\times$ $60$mm $\times$ $40$mm, which can be easily placed under the footprint of the four delta robots above. The electronics also provides a distributed control framework, with all of the low-level control being performed within each module. 

\subsection{Communication Across the Array}
To efficiently control the entire array of 64 robots, communication factors like latency, noise, and amplitude of signal need to have stable optimal values. Instead of using TTL communication using wires, which results in exhaustive cable management and noise, we use off-the-shelf ESP-01 WiFi modules operating at 115,200 baud rate, enabling low latency, low noise, and speedy wireless communication. A high-level flowchart of communication is shown in (Fig. \ref{fig:fig4}). We use protocol buffers (protobuf) to transmit control data because of their high compressibility and effectiveness in networked communications as shown in \cite{protobuf}.

%To command the delta arrays, Python-based control algorithms send desired joint positions to the Feather M0. The commands from Python are sent serially using a CP2102 (RS232-to-TTL converter) UART module. Each of the $2\times2$ modules internally runs PID control for position across the 4 linear delta robots via the  bus. 

\begin{figure}
    \centering
    \captionsetup{width=.495\textwidth}
    \includegraphics[width=0.495\textwidth]{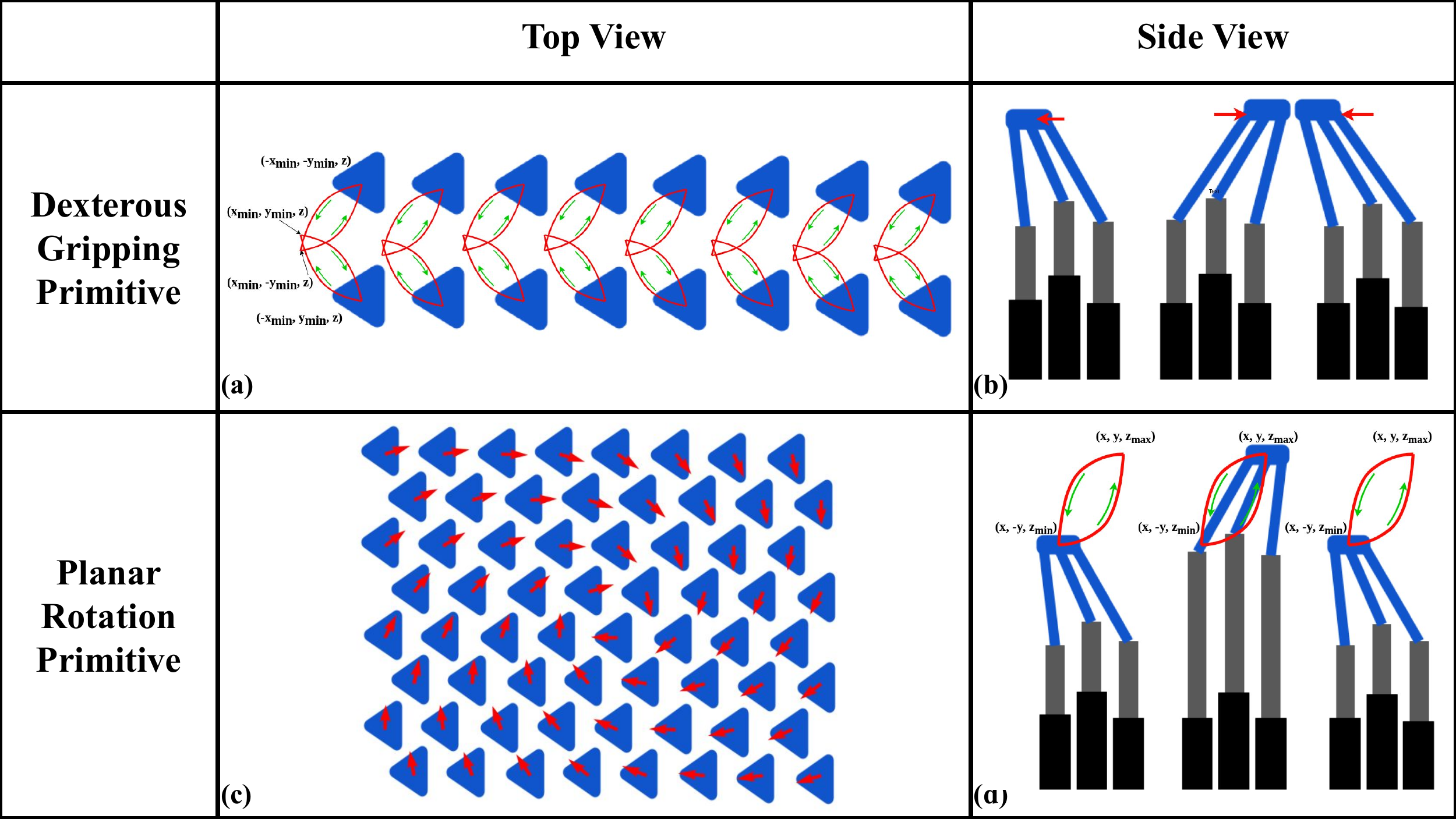}
    \caption{Top-view and side-view representation of the ellipsoid trajectories of the two-beat gaits. (a) represents 2 odd rows from the array for execution of the Dexterous Gripping Primitive and (c) represents the entire delta array executing the Planar Rotation Primitive. Both follow two-beat gait patterns shown in (b) and (d) respectively}
    \label{fig: upsideUp}
\end{figure}

\section{Predefined Distributed Manipulation Strategies}\label{strats}
The $8\times8$ array in "facing-up" mode can execute a variety of dexterous manipulation strategies distributed across its delta robots. These strategies include planar manipulations like translation, rotation, and convergence, as well as out-of-plane and prehensile manipulations. To test the capabilities of the delta array, we implemented a series of basic manipulation policies. Each delta robot is given $(x,y)$ coordinates representing its position $\vec{p}$. For every primitive, we use simple linear algebraic operations to determine the position-controlled trajectories of the delta robots. 

% We also show the capability of the delta arrays for tasks like grasping objects with multiple end effectors for dexterous manipulation. The following sections describe the respective control policies.

% \subsection{Two-beat Gait}

The delta array policies are designed as two-beat finger-gaiting strategies that repeatedly cause the deltas to make and break contact with the objects being manipulated. The planar trajectory moves in the vertical direction, with a constant gait, and along a horizontal direction as given by a high-level primitive. The movements can be considered as going from $[-\vec{p}, z_{min}]$ to $[\vec{p}, z_{max}]$ as shown in Fig.\ref{fig: upsideUp}A, where $z_{min}=7 cm$ and $z_{max}=10 cm$ are the alternating end-effector positions on $Z-axis$.
The two-beat gait means that half of the deltas in the array will be in an up configuration while the other half are in a down configuration, i.e., $180$ degrees out of phase. We use a two-beat gait to maximize the number of deltas in contact with the object at a given time as described in \cite{thompson2021towards}.

\subsection{Dexterous Gripping Primitive}
\begin{figure}
\centering
    \captionsetup{width=\columnwidth}
    \includegraphics[width=\columnwidth]{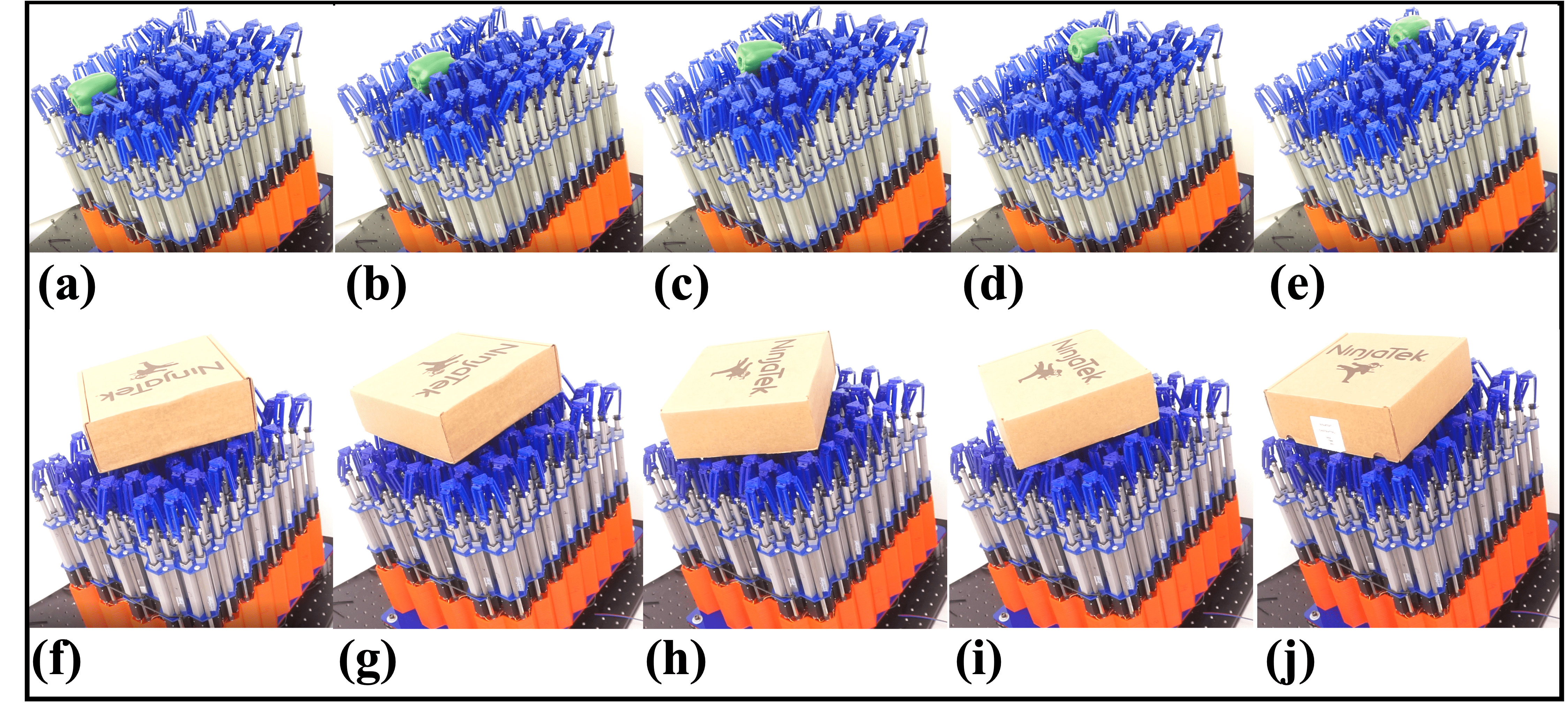}
\caption{The rows of images demonstrate manipulations of different objects using the $8\times 8$ delta array. The numbers beneath each row indicate the timestamp. (A) A toy bell pepper object that weighs $4g$ with a characteristic length of $60mm$ is transported from one edge of the array to the other using a dexterous gripping primitive from (a) to (e). (B)A box object is rotated while the position on the array stayed the same as shown from (f) to (j).}
\label{fig:fig6}
\end{figure}

Apart from purely planar manipulation strategies, we present a \emph{grip-and-push} primitive that can be deployed to grasp objects within a line of delta robots and push the object forward or backward along the line. We use a two-beat finger gait with a constant $Z$ value, and periodically switching $Y_{max}$ and $Y_{min}$ to push objects along the $X-axis$. A demonstration of the strategy on a foam bell pepper is shown in Fig.\ref{fig: upsideUp}[(a), (b)] and Fig. \ref{fig:fig6}A.

\subsection{Planar Translation Primitive}
For planar translations, a straightforward implementation of up, down, left, and right movements can be shown by placing a point along the $X$ and $Y$ axes at infinity, computing the unit distance vector from the center of the robot to that point and plug the unit vector in the aforementioned two-beat finger-gaiting pattern for planar translation of objects on the surface of the linear delta arrays. 

\subsection{Planar Rotation Primitive}
For planar rotation, the distance vector from each robot to the center is multiplied by the rotation matrix to generate rotating unit vectors for planar rotation as shown in Fig.\ref{fig: upsideUp}[(c), (d)] and Fig.\ref{fig:fig6}B. %**INSERT VECTOR Rotation TO PT FIGURE**. 

\subsection{Wall Primitive}
A unique feature of linear delta arrays is the ability to use a subset of delta robots to form walls of various shapes for restricting the movements of objects. Dexterous tasks like clamping or aligning an object along the wall and turning it around for inspection can be performed using simple yet effective policies, an inverted version of which we present in the next section.

\begin{figure}
    \centering
    \captionsetup{width=\columnwidth}
    \includegraphics[width=\columnwidth]{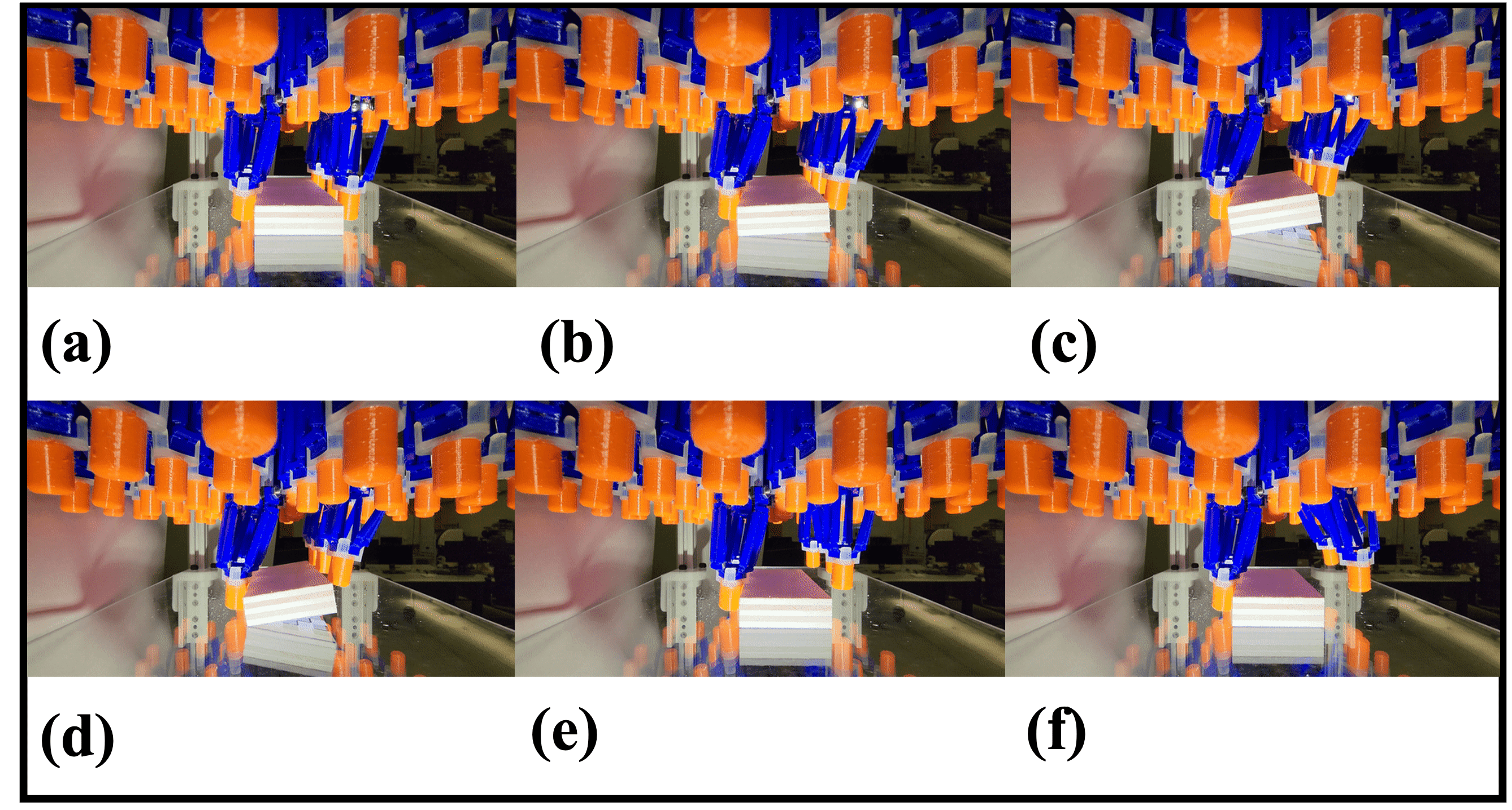}
    \caption{Using the delta robots for tilting an object using learned trajectories}
    \label{fig:object_tilt}
\end{figure}

\section{Learning Dexterous Manipulation Strategies}\label{dexmanip}
The array can also be used to learn basic dexterous manipulation skills. We design an RL environment on the real hardware with the delta array in the facing-down mode \ref{fig:object_tilt}. From the robot's camera, we track the SE(3) pose of a checkerboard pattern attached to a wooden block that the robot should manipulate using the neighboring 6 delta robots. The pose of the object is tracked to generate the reward for performing the task. We compute the L2 error between the current pose and desired pose and compare it with a threshold of 0.1 cm for translation and 0.5 rad for rotation. We use the following formula to compute the reward to maximize:
\begin{align}
    f(x)=
        \begin{cases}
            -1*T_e -3*R_e ,& \text{if } T_e > 0.1 \lor R_e > 0.5\\
            +10,              & \text{otherwise}
        \end{cases}
\end{align}
Where $T_e$ is the translational error and $R_e$ is the rotational error.

The errors generated by the vision system are used to train fingertip trajectories for grasping and tilting the wooden block using episodic relative entropy policy search (eREPS) \cite{REPS}. The trajectories are generated by weighting the output of eREPS on 5 basis functions of the DMPs. eREPS is a model-free RL algorithm that iteratively optimizes a Gaussian distribution over the weights of the DMP. We initialize the distribution with a zero mean and a diagonal covariance matrix of $0.7$.%To initialize the distribution, we define  Bézier curves as demonstration and use linear regression to obtain the corresponding DMP parameters for the mean.

In each execution episode, the robot samples the weights of the DMP, generates and executes the corresponding fingertip trajectory, and computes the resulting reward. The Gaussian policy is updated every ten episodes. The algorithm runs until a success rate of at least 90\% has been achieved, which takes approximately 270 epochs.

%At every iteration in an episode, a Bézier curve is used to generate a dynamically feasible 20-step trajectory using a DMP, generate cost associated with the deployment of the trajectory in the real world, and feedback the cost to improve the weights for eREPS. The algorithm runs until an entire episode is solved successfully at least 90\% of the time. 

\section{Experiments}\label{exps}
This section describes the experimental evaluations performed using the delta array.

%\begin{figure*}
%\includegraphics[width=\textwidth]{figures/Slide1.png}
%\end{figure*}

%\subsection{Results}\label{exps}
\subsection{Facing-up Manipulation Experiment}
We constructed an $8\times8$ delta array using the design described in Sections \ref{design} and \ref{array}. We implemented the distributed manipulation strategies explained in Section \ref{strats} for the upward facing configuration. The robot could then manipulate objects placed on top of the array. The robot successfully performed  non-prehensile translation and rotation manipulations of objects of dimensions ranging from $60$mm $\times$ $40$mm $\times$ $20$ mm to $300$mm $\times$ $300$mm $\times$ $80$mm and weights ranging from $4$g to $1$kg. Examples of different manipulations are shown in Fig. \ref{fig:fig6}. Each picture represents a snapshot of the manipulation task being performed. Our methods work on objects with a stable center of mass and a moderate coefficient of friction. Objects like soda cans with a shifting center of mass are hard to manipulate in an open-loop control setting.

\begin{figure}
    \centering
    \captionsetup{width=\columnwidth}
    \includegraphics[width=\columnwidth]{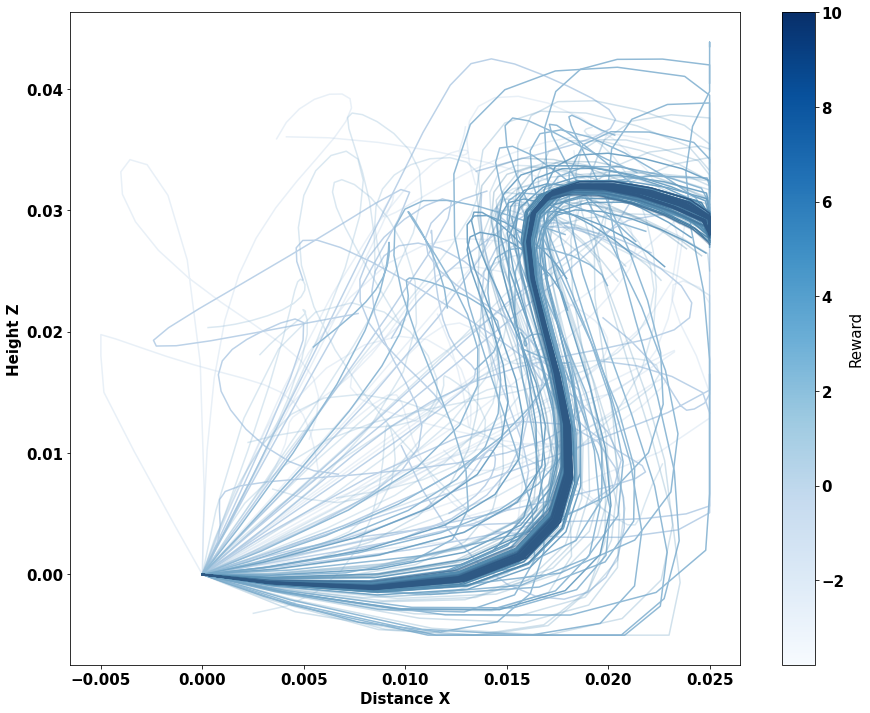}
    \caption{Using REPS to generate weights of the basis functions of a DMP to generate a 2D trajectory (curved lines from (0,0) till end-point predicted by REPS) for the delta robots. Faint trajectories represent initial exploration trajectories with low/negative rewards, darker trajectories represent learned exploitation with high rewards}
    \label{fig:graph}
\end{figure}
\subsection{Facing-down Manipulation Experiment}
We implemented the learning of dexterous manipulation strategies in Section \ref{dexmanip} for the downward facing configuration. This was done only in the facing-down configuration due to the ease of resetting the environment as compared to the upright configuration.The robot could learn to grasp and tilt objects placed on a planar surface directly in the real world without simulation. The face-down mode provides a more stable environment for easier resetting of objects between episodes. Throughout the training, the trajectories generated by the robot are shown in Fig. \ref{fig:graph}. 

In the initial stages, the algorithm explores the action spaces while generating very low rewards shown as the faint tinted trajectories in Fig. \ref{fig:graph}. As the training progresses, the actions converge to more optimal values and the model becomes exploitative to generate maximum reward consistently towards the end of the training as shown by the darker lines in Fig. \ref{fig:graph}.  The learning approach could be used to generate DMPs for other shapes of objects as well. %The learned grasping motions are object agnostic, with the challenge being able to generate appropriate reward functions for different objects. 
Due to the soft linkages, heavy objects with smooth surfaces are hard to lift using the delta robots.

\subsection{Discussion}
For the upward facing configuration, the results show that the delta arrays can be used for a variety of manipulation types. In-plane translation and rotations perform better when applied to larger objects where more delta robots make contact with the  objects at any time. In some cases, smaller objects can get stuck between deltas in the array, but the compliance keeps the system safe in these situations. The weight of objects plays an important role in the performance of the non-prehensile manipulations as well. We found that heavier objects tended to be manipulated more easily. Part of this may be due to the correlation in size and higher friction forces between the delta robots and the object. 

The wall policy allows the delta array to successfully align objects against the side of sets of delta robots. In this manner, the delta array can remove some of the uncertainty of the object's position. The wall policy can also be seen as a hybrid policy that combines the use of the translational policy with using some fingers as fixtures/obstacles. The delta array thus presents a suitable base for exploring a variety of mixed manipulation strategies in the future. %However, setting up a reset module for the upright configuration would need development of non-trivial contraptions, thus making RL policies hard to train. 

For the downward facing configuration, we observe that the design of the fingertips played an important role since higher friction generated by the fingertip surface made objects easier to tilt against the walled fingertips. The hollow cavity in the fingertips helps the surface to conform to the surface of the object to make lifting the objects easier. The compliance of the robots also adds robustness to the policy which makes it slightly more sample efficient.

\section{Conclusion}
We proposed delta arrays as a new type of dexterous manipulation robot. We presented the assembly of individual delta robot modules for close packing and overlapping workspaces, and we proposed a modular design and distributed control framework. We constructed and tested an $8\times 8$ array of delta robots and showed that the delta arrays can be used to perform a variety of manipulation strategies in two different configurations. For the facing-up configuration, our current manipulation primitives tend to be better suited for larger objects, where the redundancy of the array provides robustness. While for the facing-down configuration, we were able to learn grasping and tilting strategies directly on the hardware with low-level visual supervision. 

To show generalizability across objects, we plan to incorporate vision feedback to extract object poses and learn grasping strategies to generalize to objects of various shapes and sizes. We also plan to extend the use of the delta arrays for deformable object manipulation and to demonstrate the effectiveness of compliant multi-agent manipulation systems.

\section{Acknowledgements}
The research presented in this work was funded by the
National Science Foundation under project Grant No. CMMI-2024794 %, awarded to Professor Zeynep Temel and Professor Oliver Kroemer,  
 and Meta AI Research under award number RPS6.

%\nocite{*}  % Without this, cite articles in text using \cite{...}
\bibliographystyle{IEEEtran}
\bibliography{./IEEEfull,refs}

\end{document}